\documentclass{article}
\usepackage{arxiv}

\usepackage[utf8]{inputenc}
\usepackage[T1]{fontenc}    
\usepackage{url}            
\usepackage{amsfonts}       
\usepackage{microtype}      
\usepackage{authblk}
\usepackage{graphicx}
\graphicspath{{./images/}}
\usepackage{subcaption}
\usepackage{amsmath}
\usepackage{booktabs}
\usepackage{array}
\usepackage{tabularx}
\usepackage{multirow}
\usepackage{hyperref}
\usepackage{anyfontsize}
\usepackage{colortbl}
\usepackage{xcolor}
\usepackage{float}
\usepackage{enumitem}
\usepackage{algorithm}
\usepackage{algorithmic}

\newcommand{\PromptTemplateBox}[2]{%
{\par\vspace{0.6em}%
\begin{center}%
\setlength{\fboxsep}{6pt}%
\fcolorbox{gray!60}{gray!4}{%
\begin{minipage}{0.97\linewidth}%
\colorbox{gray!20}{\parbox{\dimexpr\linewidth-2\fboxsep\relax}{\textbf{#1}}}%
\par\vspace{0.6em}\hrule\vspace{0.8em}%
#2%
\end{minipage}}%
\end{center}%
\par\vspace{0.2em}}%
}

\providecommand{\Require}{\REQUIRE}
\providecommand{\Ensure}{\ENSURE}
\providecommand{\State}{\STATE}
\providecommand{\For}{\FOR}
\providecommand{\EndFor}{\ENDFOR}

\title{Cognitively Layered Data Synthesis for Domain Adaptation of LLMs to Space Situational Awareness}

\author[1]{Ding Linghu}
\author[1]{Cheng Wang}
\author[1]{Da Fan}
\author[1]{Wei Shi}
\author[2]{Kaifeng Yin}
\author[1]{Xiaoliang Xue}
\author[2]{Fan Yang}
\author[3]{Haiyi Ren}
\author[1]{Cong Zhang}

\affil[1]{Qian Xuesen Laboratory of Space Technology, China Academy of Space Technology, Beijing, China}
\affil[2]{China Academy of Space Technology, Beijing, China}
\affil[3]{State Key Laboratory of Space Information System and Integrated Application, Beijing, China}

\date{}

\begin{document}

\maketitle

\begin{abstract}
Large language models (LLMs) demonstrate exceptional performance on general-purpose tasks. However, transferring them to complex engineering domains such as space situational awareness (SSA) remains challenging owing to insufficient structural alignment with mission chains, the absence of higher-order cognitive supervision, and poor correspondence between data quality criteria and engineering specifications. The core bottleneck is the construction of high-quality supervised fine-tuning (SFT) datasets. To this end, we propose BD-FDG (Bloom's Taxonomy-based Domain-specific Fine-tuning Data Generation), a framework that addresses incomplete knowledge coverage, shallow cognitive depth, and limited quality controllability through three synergistic mechanisms: mission-chain-driven knowledge organization, domain-operationalized cognitive question modeling, and engineering-specification-aligned quality control. The framework uses a knowledge tree to ensure structured corpus coverage, designs a question generation scheme spanning nine domain-tailored categories and six cognitive levels from Remember to Create to produce samples with a continuous increase in difficulty, and applies a multidimensional scoring pipeline to enforce domain rigor and consistency. Using BD-FDG, we construct SSA-SFT, a domain dataset of approximately 230K samples, and fine-tune Qwen3-8B to obtain SSA-LLM-8B. Experiments show that SSA-LLM-8B achieves relative BLEU-1 improvements of 144\% (no-think) and 176\% (think) on the domain test set and a win rate of 82.21\% over the baseline in arena comparisons, while largely preserving general benchmark performance (MMLU-Pro, MATH-500). These results validate that coupling cognitive layering with structured domain knowledge and engineering-aligned quality control constitutes an effective paradigm for domain-specific SFT data construction and offers a transferable framework for adapting LLMs to complex engineering fields.
\end{abstract}


\section{Introduction}\label{sec1}

In recent years, large language models (LLMs) such as GPT-4, Llama3, Qwen3, and DeepSeek have achieved strong performance in mathematical reasoning, code generation, and general knowledge comprehension\cite{yang2025qwen3,guo2025deepseek,grattafiori2024llama}. However, a gap remains between pretrained foundation models and practical deployment; the post training stage, in particular supervised fine-tuning (SFT) and preference alignment, is essential for transforming general linguistic competence into controllable capability for task execution\cite{ouyang2022training}. Among these stages, the quality and design of SFT data directly determine downstream effectiveness in vertical domains and remain a key challenge\cite{zhao2023survey}.

SFT employs question-and-answer pairs or multiturn dialogues as its basic sample unit, providing models with directly learnable supervision signals that shift them from general language modeling toward controllable instruction following and task execution\cite{sanh2022multitask,wei2022finetuned,chung2022scaling,longpre2023flan,zhang2025instruction}. Prior work has shown that the coverage, difficulty distribution, and annotation consistency of SFT data largely determine the upper bound of model usability in downstream applications\cite{zhao2023survey,touvron2023llama2}. Importantly, SFT and preference alignment (e.g., RLHF) differ fundamentally in data modality and optimization objective: the former relies on explicit reference answers for behavior cloning, whereas the latter optimizes output distributions via pairwise preferences or reward signals\cite{ouyang2022training,bai2022rlhf}. This paper targets dataset construction at the SFT stage.

Motivated by these observations, this paper focuses on dataset construction as the entry point for domain capability enhancement and selects space situational awareness (SSA) as the research context. SSA missions span target detection and tracking, trajectory prediction, task allocation, and threat assessment and disposition decision-making\cite{chao2022modeling}, involving multidisciplinary knowledge from aerospace engineering, infrared physics, signal processing, and systems engineering, all subject to strict procedural specifications and engineering constraints\cite{foutter2024adapting,maranto2024llmsat}. In such scenarios, the challenge lies not in whether a model can recall isolated knowledge points, but in whether it can produce executable, verifiable, and reasoning compliant with specifications along the mission chain. Although retrieval-augmented generation (RAG) can supplement domain knowledge\cite{arslan2024survey}, it cannot substitute for supervised learning of task workflows, constraint conditions, and reasoning patterns. Constructing high-quality domain SFT datasets oriented toward mission chains is therefore essential. Such datasets must provide verifiable supervision signals to enable reliable deployment of general-purpose LLMs in SSA.

Despite their strong language understanding and generation capabilities, existing domain SFT data construction methods exhibit three deficiencies in the SSA context:

\begin{itemize}[label=\textbullet, leftmargin=1.8em, itemsep=0.55em, topsep=0.35em, parsep=0pt, partopsep=0pt]
    \item \textit{Inadequate structured coverage of domain knowledge:} Existing corpora lack organization aligned with the detection, tracking, prediction, assessment, and disposition mission chain, leaving systematic gaps at critical stages.

    \item \textit{Limited cognitive depth of supervision samples:} Publicly available corpora are dominated by factual recall and conceptual paraphrasing, with scarce verifiable derivation supervision targeting higher-order cognitive objectives such as analysis, evaluation, and decision trade-offs.

    \item \textit{Weak alignment between data quality and engineering specifications:} General-purpose quality assessment methods lack evaluation dimensions for relevance to engineering practice and technical integration, making it difficult to reconcile scaling with domain reliability.
\end{itemize}

To address these issues, we draw on Bloom's Taxonomy\cite{krathwohl2002revision}, which partitions cognitive objectives into six hierarchical levels (Remember, Understand, Apply, Analyze, Evaluate, and Create), and propose BD-FDG, a domain SFT dataset construction framework for SSA that provides scalable, quality-controllable, and verifiable supervision signals along three axes:

\begin{enumerate}[label=(\arabic*), leftmargin=1.8em, itemsep=0.55em, topsep=0.35em, parsep=0pt, partopsep=0pt]
    \item \textit{Structured knowledge coverage driven by the mission chain:} A knowledge tree is constructed using the mission chain hierarchy as its backbone; systematic corpus coverage is achieved through recursive tracing of seed literature, ensuring structured representation of every mission stage.

    \item \textit{Cognitively layered question modeling:} A question generation scheme spanning nine categories and six cognitive levels is designed so that supervision samples form a continuous difficulty gradient from Remember to Create, compensating for the deficit in cognitive depth in existing domain data.

    \item \textit{Multidimensional quality control oriented toward engineering specifications:} An automatic scoring and filtering pipeline evaluates samples through four coordinated components: \textit{Domain-Specific Evaluation}, \textit{Self-Containment Evaluation}, \textit{Structured Scoring Criteria}, and \textit{Key Deduction/Bonus Items}, explicitly aligning quality criteria with engineering constraints.
\end{enumerate}

\section{Related Work}\label{sec2}

\subsection{General-Purpose SFT Data Construction}

In the general-purpose setting, SFT data construction has matured into an engineering pipeline encompassing task definition, data generation, quality control, deduplication, and safety auditing\cite{longpre2023flan,liang2025dataflow,albalak2024survey}. Existing approaches fall into five paradigms: (i)~bootstrap synthesis, which achieves inexpensive scaling through instructions and answers generated by models\cite{wang2022selfinstruct,taori2023alpaca,xu2024magpie}; (ii)~difficulty evolution, which deepens reasoning via instruction rewriting and progressive complexity escalation\cite{xu2023wizardlm}; (iii)~multiturn dialogue expansion, which collects high-quality interactions through manual or semiautomatic means to enhance conversational ability\cite{ding2023ultrachat,kopf2023openassistant}; (iv)~distillation of reasoning traces, which uses explanation traces from teacher models to provide stronger supervision for higher-order reasoning\cite{mukherjee2023orca,guha2025openthoughts}; and (v)~data curation and approaches that prioritize quality, in which studies show that carefully curated data at small scale can improve alignment\cite{zhou2023lima,liu2024deita,li2024selfalign} and that data quality can outweigh sheer scale\cite{gunasekar2023textbooks,abdin2024phi,muennighoff2023scaling}. These efforts indicate that SFT data require not only sufficient scale but also a balance among difficulty gradient, format diversity, and quality controllability; yet how to reliably transfer such paradigms to constrained engineering domains remains an open question.

\subsection{Domain-Specific SFT Data Construction}

Compared with open-domain settings, domain-specific SFT data construction faces stricter challenges. Work in vertical domains such as finance\cite{wu2023bloomberggpt} and medicine\cite{singhal2023medpalm2} has shown that general-purpose instruction data cannot adequately cover specialized terminological systems and domain-specific reasoning patterns, and that the main difficulty of domain adaptation lies in constructing high-quality supervision data\cite{ling2025domain,balaskas2025framework}. Three challenges arise: first, domain knowledge coverage requires structured organization to avoid corpus sparsity and concept drift\cite{qiu2025method,chai2023circuitnet}; second, higher-order cognition and verifiable reasoning demand traceable derivation chains together with consistent evaluation criteria\cite{guha2025openthoughts,liu2025synlogic,yu2025cot}; and third, data quality assessment must be aligned with task workflows, engineering constraints, and metric systems to balance scaling with reliability\cite{zheng2023judging,liang2025dataflow}. Recent work has also demonstrated that integrating retrieval-augmented generation with domain knowledge graphs can effectively support LLM deployment in high-stakes operational environments such as robotic systems\cite{wang2026raglro}. Therefore, building a domain knowledge base oriented toward the task chain, generating verifiable question-and-answer pairs in a layered manner, and establishing an automated quality control pipeline are necessary steps for improving domain model capability\cite{zheng2022pretrained,zhong2024domain,xie2024efficient}.

\begin{figure*}[t]
\centerline{\includegraphics[width=\linewidth]{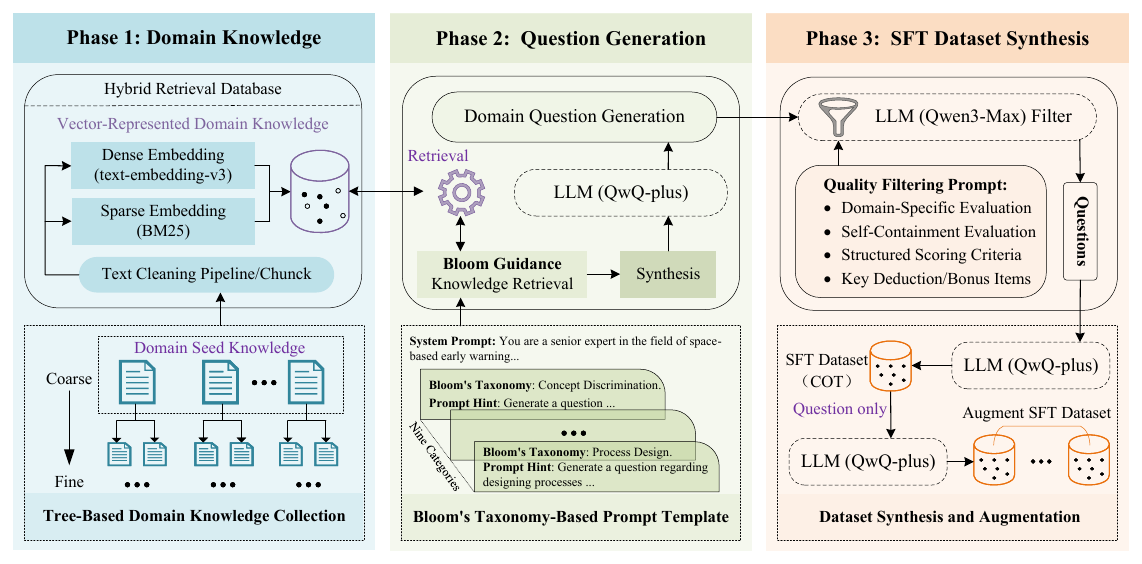}}
\caption{Architecture of the BD-FDG framework. The domain knowledge base construction stage (left) parses mission chain literature into structured chunks and builds a hybrid dense--sparse retrieval index for multisource context assembly. The question generation stage (middle) applies nine Bloom's Taxonomy-guided prompt templates across six cognitive levels to produce questions with graduated difficulty from retrieved context. The SFT data synthesis stage (right) distills reasoning traces and answers from teacher models, scores each sample along four quality dimensions, and filters by a composite threshold to yield the final SSA-SFT dataset.\label{fig:data_processing_architecture}}
\end{figure*}

\subsection{Cognitive Frameworks in Data Construction}

Bloom's Taxonomy\cite{krathwohl2002revision} classifies cognitive objectives into six hierarchical levels (Remember, Understand, Apply, Analyze, Evaluate, and Create) and has been widely adopted in educational assessment and automatic question generation. In the NLP and AI communities, this taxonomy has been used to control the cognitive complexity of generated questions\cite{toba2024large}, classify question difficulty in standardized tests, and design evaluation rubrics for LLM outputs. Recent studies have further demonstrated that Bloom-guided question generation and quiz construction with LLMs are already feasible in educational settings\cite{scaria2024automated,elkins2024teachers,duong2024bloomllm}. Although progressive difficulty escalation has proven effective for general-purpose instruction tuning\cite{xu2023wizardlm}, such approaches rely on unconstrained complexity evolution rather than a principled cognitive framework, making it difficult to ensure systematic coverage of higher-order reasoning objectives. Therefore, the novelty of our framework does not lie in applying Bloom's Taxonomy per se, but in operationalizing it for SSA-oriented SFT data construction by extending Bloom’s six cognitive levels into nine domain-specific question types aligned with the SSA mission chain. This design ensures that the generated SFT samples not only span a continuous difficulty gradient but also target specific cognitive skills relevant to SSA tasks, such as algorithm implementation, performance analysis, and solution decision-making, which are often underrepresented in existing domain datasets.

\section{Methodology}\label{sec3}

To address the three deficiencies identified in Section~\ref{sec1}, BD-FDG decomposes domain SFT data construction into three sequential stages. First, a \textit{domain knowledge base construction} stage (Section~\ref{sec3.1}) organizes mission chain literature into a structured knowledge tree and builds a hybrid retrieval index for multisource context assembly. Second, a \textit{cognitively layered question generation} stage (Section~\ref{sec3.2}) applies Bloom's Taxonomy-guided prompt templates to produce questions spanning six cognitive levels with a continuous difficulty gradient. Third, a \textit{supervised fine-tuning data synthesis} stage (Section~\ref{sec3.3}) distills reasoning traces from teacher models, scores each sample along four quality dimensions, and filters by a composite threshold to yield the final dataset. Figure~\ref{fig:data_processing_architecture} illustrates the overall architecture; the following subsections detail each stage in turn.

\subsection{Domain Knowledge Base Construction}\label{sec3.1}
We begin with foundational design literature on SSA missions and iteratively expand the collection by traversing citation links and gathering related technical materials, yielding a knowledge system with a tree structure covering system tasks, subsystems, and key technical units. This procedure focuses the corpus on critical processes and decision points in space situational awareness.

We then parse and clean the collected documents using MinerU\cite{niu2025mineru}, a vision-language document parsing tool that accurately extracts structured text, tables, and formulas from heterogeneous PDF sources. After parsing, we standardize terminology and notation and segment the text into retrievable chunks using a splitter that preserves the structure of paragraphs, tables, and code blocks. A hybrid retrieval module is built by indexing dense embeddings (\texttt{text-embedding-v3}) in Milvus alongside a BM25 sparse index, enabling retrieval that balances semantic similarity and keyword matching. For each query, the retriever first recalls $K_{\mathrm{cand}}$ candidate chunks from both indexes and reranks them using a hybrid score $Score_{\mathrm{hybrid}}=\alpha\,s_{\mathrm{dense}}+(1-\alpha)\,s_{\mathrm{BM25}}$; the top-ranked $K$ chunks are then selected as the final context for downstream question generation. Algorithm~\ref{alg:domain_knowledge_retrieval} summarizes the procedure.

\begin{algorithm}[!ht]
\caption{Domain Corpus Construction and Multiple Text Block Retrieval Process}
\label{alg:domain_knowledge_retrieval}
\begin{algorithmic}[1]\rmfamily
\Require Hybrid weight $\alpha$; candidates $K_{\mathrm{cand}}$; final Top-$K$
\Ensure For each anchor block/query, Top-$K$ relevant text blocks
\State Parse each document with MinerU; extract structured text; segment into chunks $\{c_i\}$ via splitting that preserves structure.
\State Embed each $c_i$ with \texttt{text-embedding-v3} to obtain a 1024-d vector $\mathbf{e}_i$; build a Milvus dense index and a BM25 sparse index.
\For{each anchor block or query $q$}
\State Construct query text and embedding $\mathbf{e}_q$.
\State Retrieve $K_{\mathrm{cand}}$ candidates via dense and BM25 search; rerank by $Score_{\mathrm{hybrid}}$.
\State Select Top-$K$ blocks; concatenate with the anchor to form a multisource context for question generation and distillation.
\EndFor
\end{algorithmic}
\end{algorithm}

\begin{table*}[!ht]
\centering
\caption{Question type system guided by Bloom's Taxonomy (Q1--Q9) for the space situational awareness domain.\label{tab:question_types}}
\begin{tabular*}{\textwidth}{@{\extracolsep\fill}lllp{18em}@{\extracolsep\fill}}
\toprule
\textbf{No.} & \textbf{Type Name} & \textbf{Cognitive Level} & \textbf{Assessment Objective} \\
\midrule
Q1 & Concept Discrimination & Remember/Understand & Understanding and distinguishing core concepts \\
Q2 & Principle Explanation & Understand & In-depth understanding and explanation of technical principles \\
Q3 & Formula Derivation & Understand/Apply & Mathematical modeling and theoretical analysis \\
Q4 & Parameter Calculation & Apply & Design and calculation of key parameters \\
Q5 & Algorithm Implementation & Apply/Analyze & Algorithm design and optimization \\
Q6 & Performance Analysis & Analyze & System performance analysis and comparison \\
Q7 & Process Design & Analyze/Create & System process design and task planning \\
Q8 & Solution Decision-Making & Evaluate & Engineering decision-making and solution selection \\
Q9 & Comprehensive Evaluation & Evaluate/Create & Multidimensional comprehensive evaluation and systems thinking \\
\bottomrule
\end{tabular*}
\end{table*}

\subsection{Question Generation Guided by Bloom's Taxonomy}\label{sec3.2}
Guided by Bloom's Taxonomy, we design nine prompt templates spanning six cognitive levels: Remember, Understand, Apply, Analyze, Evaluate, and Create. Each question type is defined in terms of cognitive objective, reasoning depth, and engineering applicability, ensuring coverage from basic concept recall to comprehensive evaluation. Table~\ref{tab:question_types} summarizes the nine question types guided by Bloom's Taxonomy; the full prompt templates are provided in Appendix~\ref{app:bd-fdg-prompts}.

For each anchor text block, we retrieve Top-$K$ related blocks using the hybrid dense and sparse retriever to construct a multisource context window. This context is provided to QWQ-Plus to generate questions, reasoning traces (think mode), and final answers, yielding candidate training samples that encourage synthesis across documents and structured reasoning.

\subsection{Supervised Fine-Tuning Data Synthesis}\label{sec3.3}
During data synthesis, the question, reasoning trace, and final answer generated by QWQ-Plus are organized into candidate SFT samples for the student model. Qwen-Max then performs answer quality filtering using a four-part rubric: (i)~\textit{Domain-Specific Evaluation}, which verifies whether a sample is technically sound and aligned with SSA terminology, mission workflows, and engineering constraints; (ii)~\textit{Self-Containment Evaluation}, which checks whether the answer is sufficiently complete and interpretable on its own, without relying on omitted context; (iii)~\textit{Structured Scoring Criteria}, which apply a standardized evaluation template to assess completeness, logical coherence, and internal consistency; and (iv)~\textit{Key Deduction/Bonus Items}, which introduce explicit penalties for factual errors, logical contradictions, or format violations, while allowing score adjustments for particularly rigorous and well-grounded responses.

To expand scale while preserving diversity, we perform multiple distillation of each question: each filtered question is distilled 16 times (X16) with its associated multisource context. This procedure yields complementary reasoning paths and explanations that improve coverage of complex cognitive tasks while mitigating single-path bias.
Beyond improving response diversity, the X16 multidistillation strategy also rapidly increases the effective scale of the domain-specific SFT corpus, which is particularly valuable in vertical scenarios where high-quality supervision is inherently scarce. The resulting corpus size provides greater flexibility for subsequent mixed-data fine-tuning, allowing domain-specific samples to be combined with general-purpose instruction data at well-calibrated proportions during training.

In summary, BD-FDG integrates corpus organization driven by the mission chain, question modeling guided by Bloom's Taxonomy, and multidimensional quality control to generate scalable and verifiable SFT supervision for space situational awareness.

\begin{figure*}[htb]
\centerline{\includegraphics[width=\linewidth]{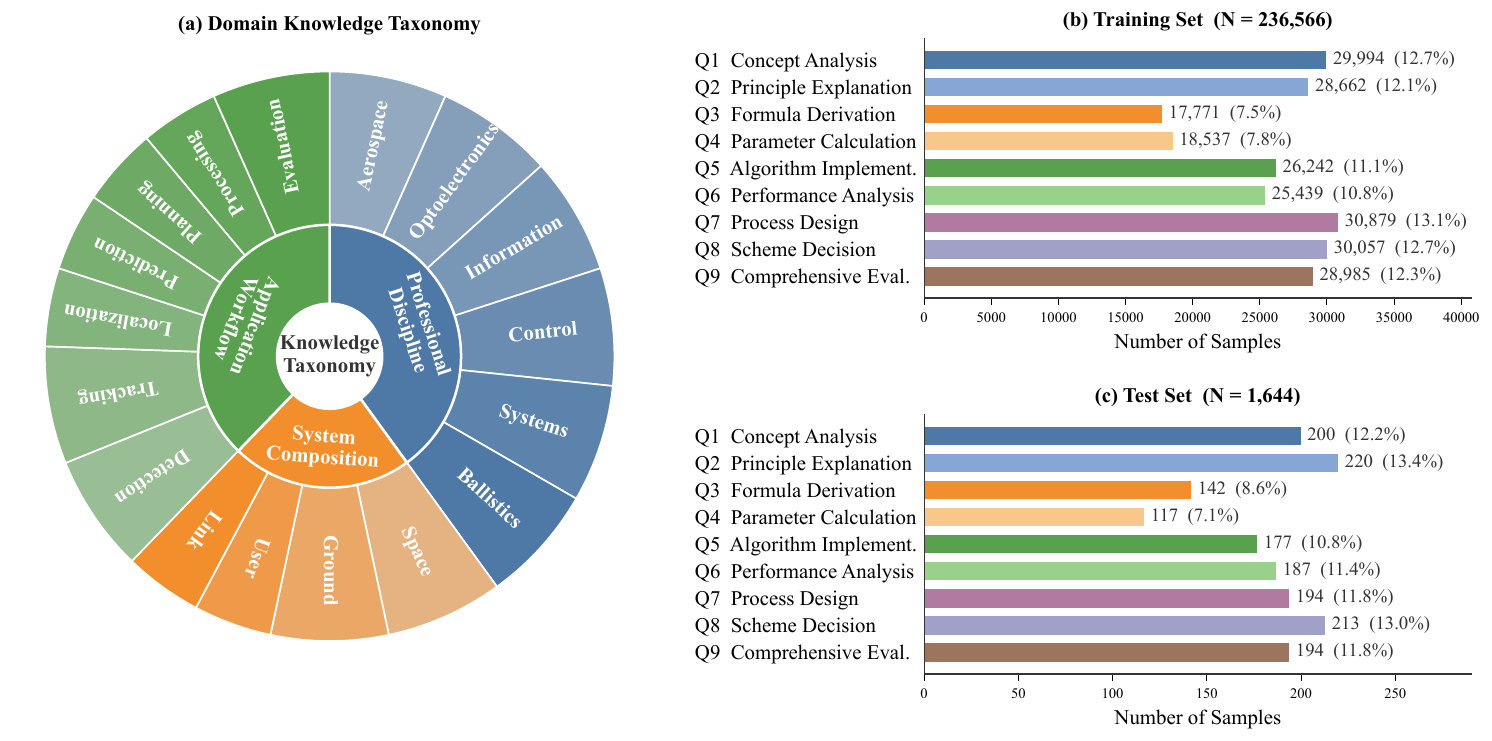}}
\caption{Overview of the SSA-SFT and SSA-Test datasets. Left: three-tier mission chain knowledge system. Right: distribution of question types and sample counts.\label{fig:sew_sft_dataset_overview}}
\end{figure*}

\subsection{SSA-SFT and SSA-Test Datasets}
We curate publicly accessible literature and technical documents to build a domain knowledge base organized by a three-tier mission chain knowledge system covering system-level tasks, subsystem modules, and key technical units. This knowledge system is specifically oriented toward representative space situational awareness application scenarios, including space debris tracking, multi-target mission planning, trajectory prediction, and threat assessment and disposition. Its three-tier mission chain structure is visualized in the left panel of Fig.~\ref{fig:sew_sft_dataset_overview}, which illustrates how domain knowledge is organized from mission-level tasks to subsystem functions and key technical units. Using BD-FDG, we generate questions, answers, and reasoning traces via QWQ-Plus, and then use Qwen-Max to score and filter the generated answers before forming SFT samples. Each question is distilled 16 times, yielding SSA-SFT, a space situational awareness SFT dataset comprising approximately 230K high-quality samples that span all nine question categories and six cognitive levels, with higher-order types (Q5 to Q9) accounting for approximately 60\% of the total to ensure adequate supervision of analysis, evaluation, and design tasks.

For evaluation, we independently construct SSA-Test using the same pipeline but with nonoverlapping source documents to prevent data leakage. SSA-Test contains 1{,}644 high-quality question and answer samples spanning the nine categories guided by Bloom's Taxonomy. To further mitigate leakage risk, we apply semantic deduplication by computing pairwise cosine similarity of question embeddings (\texttt{text-embedding-v3}) between SSA-SFT and SSA-Test, removing any test sample whose maximum similarity to the training set exceeds a threshold.

SSA-Test is used for domain question answering evaluation via BLEU/ROUGE as well as Arena Battle comparisons that assess answer quality oriented toward engineering (e.g., professionalism, completeness, and usability). Figure~\ref{fig:sew_sft_dataset_overview} summarizes the statistics of SSA-SFT and SSA-Test.

\section{Experiments}\label{sec4}

We compare the Qwen3-8B baseline with its counterpart fine-tuned on domain data SSA-LLM-8B under identical hardware and evaluation settings. Performance is assessed on the SSA-Test domain benchmark and several general benchmarks. Two inference modes (no-think vs.\ think) are analyzed to quantify the impact of explicit reasoning traces at inference time.

\subsection{Experimental Setup}

\textbf{Training and Test Datasets.} We use Qwen3-8B as the base model for supervised fine-tuning, aiming to enhance SSA competence while preserving general capabilities. The training data combines 600K general samples randomly selected from OpenThoughts3\cite{guha2025openthoughts} with approximately 230K SSA-SFT domain samples. Domain performance is evaluated on SSA-Test; general capabilities are evaluated on mathematics benchmarks (AIME24, AIME25, MATH-500), knowledge/exam benchmarks (MMLU-Pro, MMLU-Redux, C-Eval, iQuiz, GPQA-Diamond), instruction following (IFEval), and code generation (LiveCodeBench). All evaluations are conducted using the evalscope\cite{evalscope_2024} platform.

\textbf{Inference Modes and Evaluation Pipeline.} To investigate the influence of chain-of-thought reasoning on model performance, we define two inference modes: in ``no-think'' mode the model outputs only the final answer, whereas in ``think'' mode it first produces intermediate reasoning steps before the final answer. To ensure comparability, all assessments extract and compare only the final answer segment. Domain knowledge tasks are evaluated using textual overlap metrics (BLEU-1/2/3/4 and ROUGE-1/2/L-F); general benchmark tasks use standard accuracy or task-specific scores, with results reported as microaverages. The original Qwen3-8B model without domain fine-tuning serves as the baseline in all experiments.

\textbf{Training Configuration.} All experiments employ Qwen3-8B\cite{yang2025qwen3} for supervised fine-tuning. Training is executed on 8$\times$NVIDIA GPU 80\,GB GPUs using the ms-swift framework\cite{zhao2025swift} with tensor parallelism of 4, a global batch size of 32, and a maximum sequence length of 8{,}192 tokens. We use the AdamW optimizer with a weight decay of 0.01 and gradient clipping at a maximum norm of 1.0. The learning rate is set to $1\times10^{-5}$ with a linear warm-up over the first 3\% of training steps followed by cosine decay scheduling over 5 epochs.

\subsection{Main Results}

Table~\ref{tab:main_results} presents microaverage scores for Qwen3-8B and SSA-LLM-8B in both inference modes, illustrating the trade-off between domain enhancement and general capability retention.

\begin{table*}[!ht]
\centering
\caption{Test results of SSA-LLM-8B and Qwen3-8B on various test sets. All values are in \%. \textbf{Bold} indicates the better or tied best value under the same mode.\label{tab:main_results}}
\begin{tabular*}{\textwidth}{@{\extracolsep\fill}llcccc@{\extracolsep\fill}}
\toprule
\multirow{2}{*}{\textbf{Dataset}} & \multirow{2}{*}{\textbf{Task Type}} & \multicolumn{2}{c}{\textbf{no-think}} & \multicolumn{2}{c}{\textbf{think}} \\
\cmidrule{3-4} \cmidrule{5-6}
& & Qwen3-8B & SSA-LLM-8B & Qwen3-8B & SSA-LLM-8B \\
\midrule
SSA-Test        & Knowledge    & 21.33 & \textbf{52.08} & 20.75 & \textbf{57.23} \\
AIME24          & Math        & \textbf{33.33} & 23.33 & \textbf{66.67} & \textbf{66.67} \\
AIME25          & Math        & 16.67 & \textbf{23.33} & \textbf{60.00} & 56.67 \\
C-Eval          & Exam        & \textbf{79.35} & 78.01 & \textbf{83.28} & 79.20 \\
GPQA Diamond    & Knowledge    & \textbf{51.52} & 44.44 & \textbf{60.10} & \textbf{60.10} \\
IFEval          & Instruction & \textbf{84.07} & 79.66 & \textbf{85.32} & 75.89 \\
iQuiz           & Exam        & \textbf{59.17} & 57.50 & \textbf{66.67} & 60.83 \\
LiveCodeBench   & Code        & \textbf{23.08} & 19.78 & \textbf{46.70} & 40.11 \\
MATH-500        & Math        & 84.60 & \textbf{85.40} & \textbf{94.80} & \textbf{94.80} \\
MMLU-Pro        & Exam        & 64.24 & \textbf{67.13} & \textbf{74.34} & 73.15 \\
MMLU-Redux      & Exam        & 81.12 & \textbf{83.47} & 87.42 & \textbf{87.47} \\
\bottomrule
\end{tabular*}
\end{table*}

As shown in Table~\ref{tab:main_results}, fine-tuning yields substantial improvements in domain question answering. On SSA-Test, SSA-LLM-8B achieves BLEU-1 scores of 52.08\% (no-think) and 57.23\% (think), compared with 21.33\% and 20.75\% for Qwen3-8B, corresponding to relative improvements of approximately 144\% and 176\%, respectively. These results indicate that the dataset constructed by BD-FDG injects substantial domain knowledge and improves answer organization and reasoning patterns.

On general benchmarks, mathematics performance remains stable or slightly improves: MATH-500 scores range from 84.60\% to 94.80\% across all configurations, and AIME performance is largely preserved. Knowledge and exam benchmarks reveal more nuanced trade-offs: in no-think mode, MMLU-Pro and MMLU-Redux improve by approximately 2.9 and 2.4 percentage points, respectively, whereas in think mode MMLU-Pro declines slightly and MMLU-Redux remains nearly unchanged. C-Eval, GPQA-Diamond, and iQuiz exhibit declines of 1 to 7 percentage points, indicating that domain-specific fine-tuning comes at the cost of some general abilities. Instruction following and code generation (IFEval, LiveCodeBench) also decline modestly, reflecting insufficient diversity in instructions and code coverage within the training set dominated by domain data.

Comparing inference modes, think mode consistently benefits domain performance: on SSA-Test, SSA-LLM-8B improves from 52.08\% to 57.23\% in BLEU-1 (+9.9\%), whereas Qwen3-8B shows negligible change (21.33\% vs.\ 20.75\%). This asymmetry suggests that chain-of-thought reasoning is more effective when the model already possesses internalized domain knowledge to draw upon, whereas it provides limited benefit for a model lacking the requisite domain foundations. On general benchmarks, however, think mode does not uniformly help: IFEval drops from 79.66\% to 75.89\% for SSA-LLM-8B, likely because the additional reasoning steps introduce verbosity that conflicts with strict instruction-following format requirements. This pattern indicates that the benefit of explicit reasoning is task dependent, being most pronounced for knowledge-intensive open-ended questions and least beneficial for format-constrained tasks.

Overall, BD-FDG significantly enhances domain performance while keeping core general benchmarks within an acceptable range, particularly for mathematics and comprehensive exam tasks. The modest decline in instruction and code capabilities points to a clear direction for future improvement via multitask or mixed instruction fine-tuning.

\subsection{Arena Battle Evaluation}

To assess domain knowledge from two complementary perspectives, namely relative cognitive superiority and absolute textual consistency, we employ both Arena Battle (pairwise comparison by an LLM) and textual overlap metrics (BLEU/ROUGE) on SSA-Test.

\textbf{Arena Battle protocol.} Arena Battle performs pairwise comparisons of responses from two models to the same question, with a unified judge model (Qwen3-Max) selecting the preferred response based on professionalism, completeness, and usability. Compared with n-gram overlap alone, this approach better captures factors such as domain-specific terminology usage and argumentation completeness. For each question, the judge indicates its preference (A/B/tie) with a reasoned justification; overall win rates and confidence intervals (CI) are then computed.

Table~\ref{tab:arena_battle} summarizes the win rates of SSA-LLM-8B relative to Qwen3-8B.

\begin{table}[!htb]
\centering
\caption{Arena Battle results on the SSA-Test dataset (SSA-LLM-8B vs.\ Qwen3-8B).\label{tab:arena_battle}}
\begin{tabular*}{\columnwidth}{@{\extracolsep\fill}lccc@{\extracolsep\fill}}
\toprule
\textbf{Inference Mode} & \textbf{Win Rate (\%)} & \textbf{Lower CI (\%)} & \textbf{Upper CI (\%)} \\
\midrule
no-think & 82.21 & 81.50 & 82.91 \\
think    & 73.54 & 72.84 & 74.44 \\
\bottomrule
\end{tabular*}
\end{table}

As shown in Table~\ref{tab:arena_battle}, SSA-LLM-8B surpasses Qwen3-8B in both inference modes. In no-think mode, the win rate reaches 82.21\% with a 95\% confidence interval of [81.50\%, 82.91\%], indicating that the fine-tuned model is preferred in over four out of five comparisons across the 1{,}644 test questions. In think mode, the win rate is 73.54\% [72.84\%, 74.44\%], still representing a substantial margin. The narrow width of both intervals (approximately 1.4 percentage points) confirms that the observed advantage is statistically robust rather than driven by a small subset of questions. Comparing the two modes, the win rate drops by 8.67 percentage points from no-think to think, indicating that chain-of-thought prompting allows the baseline to partially compensate for its lack of domain training by organizing its latent general knowledge more effectively at inference time. However, the fine-tuned model still maintains a win rate above 73\% even in this more competitive setting, suggesting that the domain knowledge internalized through SFT provides a persistent advantage that inference-time reasoning alone cannot fully bridge.

\subsection{Fine-Grained BLEU and ROUGE Analysis}

To complement the aggregate view, Table~\ref{tab:domain_metrics} reports fine-grained textual matching metrics (BLEU-1/2/3/4, ROUGE-1/2/L-F) for both models under both inference modes.

\begin{table*}[!ht]
\centering
\caption{Fine-grained metrics on the SSA-Test dataset. All values are in \%. \textbf{Bold} indicates the better or tied best value under the same mode.\label{tab:domain_metrics}}
\begin{tabular*}{\textwidth}{@{\extracolsep\fill}lcccc@{\extracolsep\fill}}
\toprule
\multirow{2}{*}{\textbf{Metric}} & \multicolumn{2}{c}{\textbf{no-think}} & \multicolumn{2}{c}{\textbf{think}} \\
\cmidrule{2-3} \cmidrule{4-5}
& Qwen3-8B & SSA-LLM-8B & Qwen3-8B & SSA-LLM-8B \\
\midrule
BLEU-1    & 21.33 & \textbf{52.08} & 20.75 & \textbf{57.23} \\
BLEU-2    & 12.40 & \textbf{35.26} & 12.81 & \textbf{39.53} \\
BLEU-3    & 7.80 & \textbf{25.70} & 8.54 & \textbf{29.18} \\
BLEU-4    & 4.97 & \textbf{19.26} & 5.77 & \textbf{22.16} \\
ROUGE-1-F & 42.31 & \textbf{53.03} & 45.30 & \textbf{54.62} \\
ROUGE-2-F & 16.87 & \textbf{26.31} & 18.49 & \textbf{28.16} \\
ROUGE-L-F & 23.67 & \textbf{32.62} & 24.61 & \textbf{34.30} \\
\bottomrule
\end{tabular*}
\end{table*}

As shown in Table~\ref{tab:domain_metrics}, domain-specific fine-tuning is the main factor driving improvement. In no-think mode, SSA-LLM-8B surpasses Qwen3-8B across all metrics, with especially large gains in higher-order n-grams: BLEU-4 rises from 4.97\% to 19.26\% (+14.29) and ROUGE-L-F from 23.67\% to 32.62\% (+8.95). These improvements indicate that the model better aligns with domain expression patterns in terms of phrase spans, structural matching, and answer organization, confirming the effectiveness of SFT data constructed by BD-FDG.

Explicit chain-of-thought reasoning (think mode) yields stable but more modest improvements across both models, with a larger gain for SSA-LLM-8B. For Qwen3-8B, think mode raises BLEU-4 from 4.97\% to 5.77\% and ROUGE-1-F from 42.31\% to 45.30\%, suggesting that explicit reasoning reduces content omission. For SSA-LLM-8B, think mode further lifts BLEU-4 from 19.26\% to 22.16\% and ROUGE-L-F from 32.62\% to 34.30\%, indicating that chain-of-thought reasoning is more effective when the model already possesses domain knowledge. Together, domain training signals and explicit reasoning at inference time play complementary roles: the former sets the upper bound of knowledge and expression, while the latter improves organization and consistency during generation.

\subsection{Hyperparameter Analysis of Hybrid Retrieval}

To systematically examine how key hybrid retrieval hyperparameters affect generation quality, we conduct a two-dimensional ablation over retrieval depth Top-$K$ and the weighting factor for semantics and keywords $\alpha$. Question-and-answer pairs are generated under 25 parameter combinations ($\alpha \in \{0.00, 0.25, 0.50, 0.75, 1.00\}$, $K \in \{1, 3, 5, 7, 9\}$) and evaluated by Qwen3-Max using a multidimensional scoring system: multisource knowledge integration (0 to 5), question complexity integration (0 to 3), answer integration quality (0 to 3), and a penalty term ($-2$ to $0$), yielding a total score in $[0, 10]$.

\begin{figure}[htb]
\centerline{\includegraphics[width=\columnwidth]{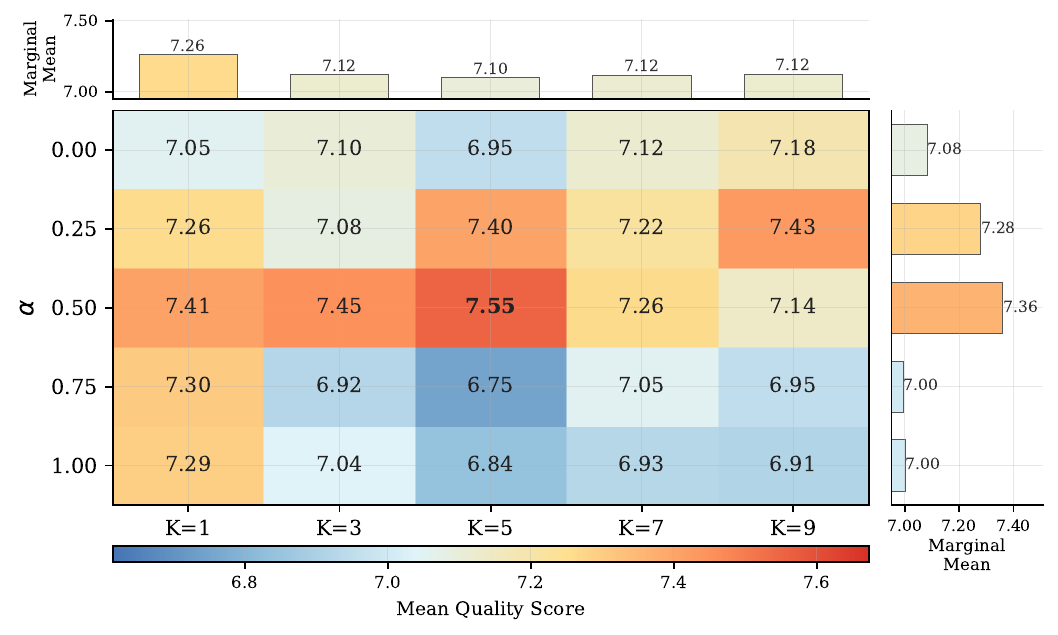}}
\caption{Two-dimensional ablation results. The central heatmap shows average quality scores for each $(\alpha, K)$ configuration; marginal bar plots depict the main effects of Top-$K$ (top) and $\alpha$ (right). The optimal configuration $(\alpha\!=\!0.50, K\!=\!5)$ attains a score of 7.55.\label{fig:2d_ablation_heatmap}}
\end{figure}

Figure~\ref{fig:2d_ablation_heatmap} presents the results. Three patterns are observed:

\textbf{(1) Inverted-U Top-$K$ effect.} With $K\!=\!1$, a single text segment offers limited integration across sources, yielding a relatively low score (7.05 at $\alpha\!=\!0.00$). As $K$ increases to 3 to 5, richer multisource input substantially improves integration quality, peaking around $K\!=\!5$. Beyond $K\!=\!7$, redundant information and semantic conflicts degrade quality, producing diminishing or negative returns. This nonmonotonic trend indicates an optimal retrieval depth range: too shallow starves the model of information; too deep introduces noise.

\textbf{(2) Optimal balance point for $\alpha$.} The parameter $\alpha$ regulates the relative weight of semantic similarity versus keyword matching ($\alpha\!=\!0$: pure keyword; $\alpha\!=\!1$: pure semantic). Results show that $\alpha\!=\!0.50$ achieves the best or second-best performance across most Top-$K$ settings, with $(\alpha\!=\!0.50, K\!=\!5)$ attaining the highest score of 7.55. In contrast, pure semantic retrieval ($\alpha\!=\!1.00$) degrades at higher $K$ (only 6.84 at $K\!=\!5$), likely due to excessive recall of semantically similar but redundant content.

\textbf{(3) Notable parameter interaction.} The marginal distributions show that the Top-$K$ main effect peaks at $K\!=\!5$ and the $\alpha$ main effect peaks at $\alpha\!=\!0.50$. An interaction is also evident: in the low-$\alpha$ regime (dominated by keywords), larger $K$ compensates by adding semantic diversity; in the high-$\alpha$ regime (dominated by semantics), moderate $K$ is essential to avoid redundancy.

Based on this analysis, we adopt $\alpha\!=\!0.50$, $K\!=\!5$ as the default configuration for dataset construction, balancing information richness with contextual controllability.

\section{Conclusion}\label{sec5}

In this paper, we address the limited domain competence of general-purpose LLMs for space situational awareness by introducing BD-FDG, a domain SFT data generation framework. While Bloom-guided question generation has been explored in educational contexts~\cite{scaria2024automated,elkins2024teachers,duong2024bloomllm}, BD-FDG extends this idea to mission-chain-driven engineering domains by coupling domain-operationalized cognitive layering with structured knowledge organization and specification-aligned quality control. Through these synergistic mechanisms together with multisource knowledge synthesis, we construct SSA-SFT, comprising approximately 230K high-quality samples.

Qwen3-8B fine-tuned on SSA-SFT consistently surpasses the baseline in arena comparisons, achieving win rates of 82.21\% (no-think) and 73.54\% (think), and substantially improves domain question answering (e.g., BLEU-1 increases from 20.75\% to 57.23\% in think mode). The model largely preserves general benchmark performance (e.g., MATH-500 and MMLU-Pro), indicating that BD-FDG improves domain capability beyond superficial memorization by providing cognitively structured supervision signals aligned with the SSA mission chain, spanning from foundational recall to integrative design.

This work has several limitations. Full parameter fine-tuning on 8$\times$GPU 80\,GB GPUs incurs substantial computational cost, constraining broader multimodel comparisons. In addition, reliance on teacher models during distillation may transfer biases into the student model, and the domain knowledge base is constructed exclusively from publicly available literature, meaning that restricted operational data covering mission-critical decision procedures are not included, potentially limiting corpus completeness at the highest operational levels. Future work will pursue three directions: (i)~incorporating human expert evaluation alongside automated metrics to provide a more reliable assessment of domain competence; (ii)~applying BD-FDG to fine-tune LLMs of different scales and architectures (e.g., Llama, DeepSeek) to further validate the framework's model-agnostic properties; and (iii)~extending the BD-FDG framework to other complex engineering domains (e.g., autonomous driving and power grid operation) to validate its generalizability as a domain adaptation paradigm.

\bibliographystyle{unsrt}
\bibliography{paper-list}

\appendix

\section{Bloom's Taxonomy-Based Prompt Templates\label{app:bd-fdg-prompts}}
\vspace*{12pt}
This appendix summarizes the Bloom's Taxonomy-grounded prompt templates used in our BD-FDG data synthesis pipeline.

\subsection{Nine Bloom-guided question types (Q1--Q9)\label{app1.1a}}

\PromptTemplateBox{Q1: Concept Discrimination}{%
\begin{itemize}[leftmargin=1.2em,noitemsep,topsep=0pt]
\item \textbf{Objective:} Accurate understanding and differentiation of core concepts.
\item \textbf{Design hint:} Ask for comparison, distinction, or definition of key professional concepts.
\item \textbf{Example prefix:} ``Distinguish between \dots and \dots''
\item \textbf{Answer guideline (AG):} Define each concept clearly; compare across multiple dimensions; link to practical use; use a table when helpful.
\end{itemize}
}

\PromptTemplateBox{Q2: Principle Explanation}{%
\begin{itemize}[leftmargin=1.2em,noitemsep,topsep=0pt]
\item \textbf{Objective:} Explain fundamental principles and mechanisms.
\item \textbf{Design hint:} Ask to explain how a system/phenomenon works or why it behaves so.
\item \textbf{Example prefix:} ``Explain the working principle of \dots''
\item \textbf{Answer guideline (AG):} Describe the core principle and mechanism; explain key physical/technical processes and causality; optionally use formulas.
\end{itemize}
}

\PromptTemplateBox{Q3: Formula Derivation}{%
\begin{itemize}[leftmargin=1.2em,noitemsep,topsep=0pt]
\item \textbf{Objective:} Mathematical modeling and derivation.
\item \textbf{Design hint:} Ask for derivation of key formulas, model formulation, or proof-like reasoning.
\item \textbf{Example prefix:} ``Derive the mathematical expression for \dots''
\item \textbf{Answer guideline (AG):} State assumptions and boundary conditions; derive step-by-step; explain variable meanings; specify validity range.
\end{itemize}
}

\PromptTemplateBox{Q4: Parameter Calculation}{%
\begin{itemize}[leftmargin=1.2em,noitemsep,topsep=0pt]
\item \textbf{Objective:} Compute and analyze specific parameters.
\item \textbf{Design hint:} Ask for numerical computation or parameter estimation/analysis.
\item \textbf{Example prefix:} ``Compute the numerical value of \dots''
\item \textbf{Answer guideline (AG):} List knowns/unknowns; choose formulas; show detailed calculations; interpret the physical meaning of results.
\end{itemize}
}

\PromptTemplateBox{Q5: Algorithm Implementation}{%
\begin{itemize}[leftmargin=1.2em,noitemsep,topsep=0pt]
\item \textbf{Objective:} Algorithm design, implementation, and optimization.
\item \textbf{Design hint:} Ask for algorithm steps, pseudocode, or implementation details.
\item \textbf{Example prefix:} ``Design an algorithmic workflow for \dots''
\item \textbf{Answer guideline (AG):} Provide the overall idea; list detailed steps; give pseudocode; analyze complexity and performance.
\end{itemize}
}

\PromptTemplateBox{Q6: Performance Analysis}{%
\begin{itemize}[leftmargin=1.2em,noitemsep,topsep=0pt]
\item \textbf{Objective:} Analyze and evaluate system performance.
\item \textbf{Design hint:} Ask about metrics, bottlenecks, pros/cons, and trade-offs.
\item \textbf{Example prefix:} ``Analyze the performance characteristics of \dots''
\item \textbf{Answer guideline (AG):} Define metrics; analyze influencing factors; compare advantages/limitations and scenarios; propose optimizations.
\end{itemize}
}

\PromptTemplateBox{Q7: Process Design}{%
\begin{itemize}[leftmargin=1.2em,noitemsep,topsep=0pt]
\item \textbf{Objective:} Engineering process planning and workflow design.
\item \textbf{Design hint:} Ask for an end-to-end workflow, processing steps, or system procedure.
\item \textbf{Example prefix:} ``Design the processing workflow for \dots''
\item \textbf{Answer guideline (AG):} Provide a complete framework; describe key steps; specify inputs/outputs; discuss exception handling.
\end{itemize}
}

\PromptTemplateBox{Q8: Solution Decision-Making}{%
\begin{itemize}[leftmargin=1.2em,noitemsep,topsep=0pt]
\item \textbf{Objective:} Integrated solution design and decision analysis.
\item \textbf{Design hint:} Ask for option comparison, trade-off reasoning, and final recommendation.
\item \textbf{Example prefix:} ``How to choose the optimal solution for \dots''
\item \textbf{Answer guideline (AG):} Analyze background and constraints; propose alternatives; build an evaluation framework; recommend with justification.
\end{itemize}
}

\PromptTemplateBox{Q9: Comprehensive Evaluation}{%
\begin{itemize}[leftmargin=1.2em,noitemsep,topsep=0pt]
\item \textbf{Objective:} Cross-aspect, system-level analysis and evaluation.
\item \textbf{Design hint:} Ask for multi-dimensional assessment, subsystem contributions, and improvement directions.
\item \textbf{Example prefix:} ``Comprehensively evaluate the overall effectiveness of \dots''
\item \textbf{Answer guideline (AG):} Build a multi-dimensional framework; analyze subsystem contributions; synthesize overall effectiveness; propose improvements.
\end{itemize}
}

\end{document}